# Robust learning Bayesian networks for prior belief

Maomi Ueno
Graduate School of Information Systems, University of Electro-Communications
1-5-1, Chofugaoka, Chofu-shi, Tokyo, 182-8585, Japan
ueno@ai.is.uec.ac.jp

## Abstract

Recent reports have described that learning Bayesian networks are highly sensitive to the chosen equivalent sample size (ESS) in the Bayesian Dirichlet equivalence uniform (BDeu). This sensitivity often engenders some unstable or undesirable results. This paper describes some asymptotic analyses of BDeu to explain the reasons for the sensitivity and its effects. Furthermore, this paper presents a proposal for a robust learning score for ESS by eliminating the sensitive factors from the approximation of log-BDeu.

## 1 Introduction

An extremely popular Bayesian network learning score is the marginal likelihood (ML) score (using a Dirichlet prior over model parameters), which finds the maximum a posteriori (MAP) structure, as described by Buntine (1991) and Heckerman *et al.* (1995). In addition, the Dirichlet prior is known as a distribution that ensures likelihood equivalence; this score is known as "Bayesian Dirichlet equivalence (BDe)" (Heckerman *et al.*, 1995). Given no prior knowledge, the Bayesian Dirichlet equivalence uniform (BDeu), as proposed earlier by Buntine (1991), is often used. Actually, BDe(u) requires an "equivalent sample size (ESS)", which reflects the degree of a user's prior belief. Moreover, recent studies have demonstrated that ESS plays an important role in the resulting network structure estimate.

Steck and Jaakkola (2002) demonstrated that the deletion of an arc in a Bayesian network is more likely to occur as ESS goes asymptotically to zero for a large sample. Actually, the learned network becomes an empty graph when ESS approaches zero.

This result is particularly surprising because it had been believed that the likelihood, which has consistency, would become dominant in the score as ESS approaches zero. That study also demonstrated that when the value of ESS becomes large, the number of arcs in the structure tends to increase to a complete graph, which is also counterintuitive because it had been believed that a Bayesian prior relaxed overfitting in learning. Then increasing ESS blocked the addition of extra arcs.

Silander, Kontkanen and Myllymaki (2007) conducted a series of empirical experiments to find the optimum ESS-value of BDeu. The results confirmed earlier results described by Steck and Jaakkola (2002), showing that the solution of the network structure is highly sensitive to the chosen ESS-value. To solve the sensitivity problem of BDeu, they proposed an empirical Bayes method of optimizing ESS to maximize the ML.

Steck (2008) showed that the log-Bayes factor of dependence between two nodes using BDeu is expressible as a tradeoff between the skewness (non-uniformity) of the sample distribution and model complexity. This result is almost identical to the Akaike information criterion (AIC; Akaike, 1974). Additionally, Akaike proposed an empirical method of optimizing ESS to minimize the expected error measured using AIC.

To clarify the properties of BDe(u), Ueno (2010) analyzed the log-BDe(u) asymptotically, finding the result that it is decomposed into the log-posterior and the penalty term of the complexity which reflects the difference between the learned structure from data and the hypothetical structure from the user's knowledge. As the two structures become equivalent, the penalty term is minimized with the fixed ESS. Con-versely, the term increases to the degree that the two structures become different. Furthermore, the result suggests that a tradeoff exists between the role of ESS in the log-posterior (which helps to block extra arcs) and its role in the penalty term (which helps to add extra arcs). That tradeoff might cause the BDeu score to be highly sensitive to ESS and might make it more difficult to determine an approximate ESS. However, Ueno (2010) was not able to identify the causes of the tradeoff in BDeu or a means to solve the problems. In addition, although he assumed implicitly that all the hyperparameters were more than 1.0, this is an impractical constraint, especially for BDeu, because the hyperparameters usually become less than 1.0 when the parameters become numerous.

Thereby, this work expands on Ueno(2010)'s approximation of BDeu by relaxing the constraint on the hyperparameters to obtain more accurate results. The results show that BDeu becomes more sensitive to ESS as ESS approaches zero. Furthermore, based on results of some asymptotic analyses, this study identifies the reasons for the sensitivity of BDeu by decomposing it into two parts: (1) the prior term that is independent of data, and (2) the likelihood term that reflects data. Results of this study show that the role of the prior term rapidly changes from strongly blocking extra arcs to strongly helping to add arcs. In fact, the results show that the prior term is highly sensitive to ESS and that it causes some odd phenomena of BDeu.

The results also show that the prior of BDeu does not represent ignorance of prior knowledge but represents a user's prior belief for the uniformity of conditional distribution, which causes the optimal ESS to become large/small when the empirical conditional distribution becomes uniform/skewed. This is the main factor that derives the sensitivity of BDeu to ESS.

Additionally, to solve the sensitivity problem, we propose a robust learning score (called "NIP-BIC ") for ESS by eliminating the sensitivity factors from the approximation of log-BDeu because it is impossible to eliminate them directly from the log-BDeu function. Numerical experiments show that NIP-BIC is effective especially when we have no prior knowledge.

## 2 Learning Bayesian networks

Let $\{x_1, x_2, \cdots, x_N\}$ be a set of $N$ discrete variables, each of which can take a value in the set of states $\{1, \cdots, r_i\}$. Here, $x_i = k$ means that an $x_i$ is state $k$. According to the Bayesian network structure $g \in G$, the joint probability distribution is given as

$$p(x_1, x_2, \cdots, x_N \mid g) = \prod_{i=1}^{N} p(x_i \mid \Pi_i, g), \quad (1)$$

where $G$ signifies the possible set of Bayesian network structures, and where $\Pi_i$ denotes the parent variable set of $x_i$.

Next, we introduce the problem of learning a Bayesian network. Let $\theta_{ijk}$ be a conditional probability parameter of $x_i = k$ when the $j$-th instance of the parents of $x_i$ is observed (We write $\Pi_i = j$). Buntine (1991) assumed the Dirichlet prior and used an expected a posteriori (EAP) estimator as the parameter estimator $\widehat{\Theta} = (\hat{\theta}_{ijk}), (i = 1, \cdots, N, j = 1, \cdots, q_i, k = 1, \cdots, r_i - 1)$:

$$\hat{\theta}_{ijk} = \frac{\alpha_{ijk} + n_{ijk}}{\alpha_{ij} + n_{ij}}, (k = 1, \cdots, r_i - 1), \quad (2)$$

where $n_{ijk}$ represents the number of samples of $x_i = k$ when $\Pi_i = j$ and $n_{ij} = \sum_{k=1}^{r_i} n_{ijk}$, and where $\alpha_{ijk}$ denotes the hyperparameters of the Dirichlet prior distributions. ($\alpha_{ijk}$ is a pseudo-sample corresponding to $n_{ijk}$), $\alpha_{ij} = \sum_{k=1}^{r_i} \alpha_{ijk}$, and $\hat{\theta}_{ijr_i} = 1 - \sum_{k=1}^{r_i - 1} \hat{\theta}_{ijk}$.

The marginal likelihood is obtained as

$$p(\mathbf{X} \mid g, \alpha) = \prod_{i=1}^{N} \prod_{j=1}^{q_i} \frac{\Gamma(\alpha_{ij})}{\Gamma(\alpha_{ij} + n_{ij})} \prod_{k=1}^{r_i} \frac{\Gamma(\alpha_{ijk} + n_{ijk})}{\Gamma(\alpha_{ijk})}. \quad (3)$$

Here, $q_i$ signifies the number of instances of $\Pi_i$ in which $q_i = \prod_{x_l \in \Pi_i} r_l$. Also, $\mathbf{X}$ is a dataset. The problem of learning a Bayesian network is to find the MAP structure which maximizes the score (3).

Particularly, Heckerman *et al.* (1995) presented a sufficient condition for satisfying the likelihood equivalence assumption in the form of the following constraint related to hyperparameters of (3):

$$\alpha_{ijk} = \alpha p(x_i = k, \Pi_i = j \mid g^h). \quad (4)$$

Here, $\alpha$ is the user-determined equivalent sample size (ESS) and $g^h$ is the hypothetical Bayesian network structure that reflects a user's prior knowledge. This metric was designated as the Bayesian Dirichlet equivalence (BDe) score metric.

As Buntine (1991) described, $\alpha_{ijk} = \frac{\alpha}{(r_i q_i)}$ is regarded as a special case of the BDe metric. Heckerman *et al.* (1995) called this special case "BDeu". $\alpha_{ijk} = \frac{\alpha}{(r_i q_i)}$ does not mean "uniform prior" but "the same value of all hyperparameters for a variable".

## 3 Previous works

For cases of which we have no prior knowledge, BDeu is often used in practice. Heckerman *et al.* (1995) reported, as a result of their comparative analyses of BDeu and BDe, that BDeu is better than BDe unless the user's beliefs are close to the true model. BDeu requires an "equivalent sample size (ESS)", which is the value of a free parameter specified by the user. Recent reports have described that ESS in BDeu plays an important role in learning Bayesian networks (Steck and Jaakkola, 2002, Silander, Kontkanen and Myllymaki, 2007) .

To clarify the mechanism of BDe, Ueno (2010) analyzed the log-BDe asymptotically and derived the following theorem.

**Theorem 1.** *(Ueno 2010) When $\alpha+n$ is sufficiently large, log-BDe converges to*

$$\log p(\boldsymbol{X} \mid g, \alpha) = \log p(\widehat{\Theta} \mid \mathbf{X}, g, \alpha) \qquad (5)$$
$$-\frac{1}{2} \sum_{i=1}^{N} \sum_{j=1}^{q_i} \sum_{k=1}^{r_i} \frac{r_i - 1}{r_i} \log\left(1 + \frac{n_{ijk}}{\alpha_{ijk}}\right) + const,$$

*where* $\log p(\widehat{\Theta} \mid \mathbf{X}, g, \alpha) =$
$$\sum_{i=1}^{N} \sum_{j=1}^{q_i} \sum_{k=1}^{r_i} (\alpha_{ijk} + n_{ijk}) \log \frac{(\alpha_{ijk} + n_{ijk})}{(\alpha_{ij} + n_{ij})},$$

and *const* is the term that is independent of the number of parameters. From (5), log-BDe can be decomposed into two factors: (1) a log-posterior term $\log p(\widehat{\Theta} \mid \mathbf{X}, g, \alpha)$ and (2) a penalty term $\frac{1}{2} \sum_{i=1}^{N} \sum_{j=1}^{q_i} \sum_{k=1}^{r_i} \frac{r_i-1}{r_i} \log\left(1 + \frac{n_{ijk}}{\alpha_{ijk}}\right)$. $\sum_{i=1}^{N} \sum_{j=1}^{q_i} \sum_{k=1}^{r_i} \frac{r_i-1}{r_i}$ is the number of parameters.

This well known model selection formula is generally interpreted (1) as reflecting the fit to the data and (2) as signifying the penalty that blocks extra arcs from being added.

Ueno (2010) described that the term $\sum_{i=1}^{N} \sum_{j=1}^{q_i} \sum_{k=1}^{r_i} \log\left(1 + \frac{n_{ijk}}{\alpha_{ijk}}\right)$ in (5) reflects the difference between the learned structure from data and the hypothetical structure $g^h$ from the user's knowledge in BDe. To the degree that the two structures are equivalent, the penalty term is minimized with the fixed ESS. Conversely, to the degree that the two structures differ, the term is larger. Moreover, from (5), $\alpha$ determines the magnitude of the user's prior belief for a hypothetical structure $g^h$. Consequently, the mechanism of BDe makes sense to us when we have approximate prior knowledge. However, Ueno (2010) did not present a sufficiently detailed discussion of the mechanism of BDeu for circumstances in which we have no knowledge.

Therefore, we expand on Theorem 1(Ueno, 2010) to derive a more accurate asymptotic analysis of BDeu and to identify the main reasons for its sensitivity to ESS.

## 4 Asymptotic analysis of BDeu

Asymptotic analyses of the log-BDeu are presented to find the reason for the sensitivity. First, from (3), we decompose log-BDeu into two parts (1) the prior term (the first term in (6)), which is independent of data and (2) the likelihood term (the second term in (6)), which reflects data:

$$\log p(\mathbf{X} \mid g) = \sum_{i=1}^{N} \sum_{j=1}^{q_i} \left(\log \Gamma(\alpha_{ij}) - \sum_{k=1}^{r_i} \log \Gamma(\alpha_{ijk})\right) \quad (6)$$
$$+ \sum_{i=1}^{N} \sum_{j=1}^{q_i} \left(\sum_{k=1}^{r_i} \log \Gamma(\alpha_{ijk} + n_{ijk}) - \log \Gamma(\alpha_{ij} + n_{ij})\right),$$

where $\alpha_{ijk} = \frac{\alpha}{r_i q_i}$. This section primarily presents analysis of (1) the prior term and (2) the likelihood term, respectively, and secondarily presents integration of the results to support the asymptotic analysis of BDeu.

### 4.1 Analysis of prior term

Silander, Kontkanen and Myllymaki (2007), based on results of some simulation experiments, reported that the prior term acts as a complexity penalizing factor and monotonically decreases concomitantly with increasing $\alpha$. This paper mathematically derives the following theorem.

**Theorem 2.** *When all the hyperparameters are less than $1.0 (\alpha < r_i q_i)$, the prior term is approximated as*

$$\sum_{i=1}^{N} \sum_{j=1}^{q_i} \left(\log \Gamma(\alpha_{ij}) - \sum_{k=1}^{r_i} \log \Gamma(\alpha_{ijk})\right)$$
$$= \sum_{i=1}^{N} q_i(r_i - 1) \log \frac{\alpha}{r_i q_i} + \mathcal{O}(1).$$

*Proof.* When $\alpha_{ijk}$ is sufficiently small, approximating $1/\Gamma(\alpha_{ijk}) = \alpha_{ijk} + \mathcal{O}(\alpha_{ijk}^2)$ (Steck and Jaakkola, 2002), we obtain

$$\sum_{i=1}^{N} \sum_{j=1}^{q_i} \left(\log \Gamma(\alpha_{ij}) - \sum_{k=1}^{r_i} \log \Gamma(\alpha_{ijk})\right)$$

$$= \sum_{i=1}^{N}\sum_{j=1}^{q_i}\left(\sum_{k=1}^{r_i}\log\alpha_{ijk} - \log\alpha_{ij}\right) + \mathcal{O}(max_{(i)}(\frac{\alpha}{r_i q_i})^2).$$

Using Jensen's inequality for $\alpha \in (0, 1)$ because the log function is a convex function,

$$\frac{1}{r_i}\sum_{k=1}^{r_i}\log\alpha_{ijk} + \log r_i \leq \log\alpha_{ij}.$$

$$\text{From } r_i \geq 1.0, \frac{1}{r_i}\sum_{k=1}^{r_i}\log\alpha_{ijk} \leq \log\alpha_{ij}. \quad (7)$$

Consequently, we obtain

$$\sum_{i=1}^{N}\sum_{j=1}^{q_i}\left(\log\Gamma(\alpha_{ij}) - \sum_{k=1}^{r_i}\log\Gamma(\alpha_{ijk})\right)$$
$$\leq \sum_{i=1}^{N}\sum_{j=1}^{q_i}\frac{(r_i-1)}{r_i}\sum_{k=1}^{r_i}\log\alpha_{ijk} + \mathcal{O}(1).$$
$$= \sum_{i=1}^{N} q_i(r_i-1)\log\frac{\alpha}{r_i q_i} + \mathcal{O}(1).$$

□

From Theorem 2, when all hyperparameters are less than 1.0, the term $\sum_{i=1}^{N} q_i(r_i-1)\log\frac{\alpha}{r_i q_i}$ dominates the prior term and increases the complexity penalty rapidly as the number of parameters increases. Consequently, BDeu produces empty graph structures when $\alpha$ approaches zero.

**Theorem 3.** *When all the hyperparameters are greater than 1.0 ($\alpha \geq r_i q_i$), the prior term can be approximated as*

$$\sum_{i=1}^{N}\sum_{j=1}^{q_i}\left(\log\Gamma(\alpha_{ij}) - \sum_{k=1}^{r_i}\log\Gamma(\alpha_{ijk})\right)$$
$$= \sum_{i=1}^{N}\alpha\log r_i + \frac{1}{2}\sum_{i=1}^{N} q_i(r_i-1)\log\frac{\alpha}{2\pi(r_i q_i)} + \mathcal{O}(1).$$

*Proof.* Here, we use the following Stirling series (c.f. Box and Tiao, 1992), as

$$\log\Gamma(a) = \frac{1}{2}\log(2\pi) + \left(a - \frac{1}{2}\right)\log a - a + \mathcal{O}\left(\frac{1}{a}\right).$$

When all the hyperparameters are greater than 1.0 ($\alpha \geq r_i q_i$), we obtain

$$\sum_{i=1}^{N}\sum_{j=1}^{q_i}\left(\log\Gamma(\alpha_{ij}) - \sum_{k=1}^{r_i}\log\Gamma(\alpha_{ijk})\right) =$$
$$-\sum_{i=1}^{N}\sum_{j=1}^{q_i}\sum_{k=1}^{r_i}\alpha_{ijk}\log\frac{\alpha_{ijk}}{\alpha_{ij}}$$
$$-\frac{1}{2}\sum_{i=1}^{N}\sum_{j=1}^{q_i}\left((r_i-1)\log(2\pi) - \sum_{k=1}^{r_i}\log\alpha_{ijk} + \log\alpha_{ij}\right)$$
$$+\mathcal{O}(max_i(\frac{r_i q_i}{\alpha})),$$

using Jensen's inequality for $\alpha \leq 1.0$, according to (7)

$$\frac{1}{r_i}\sum_{k=1}^{r_i}\log\alpha_{ijk} \geq \log\alpha_{ij}, \text{ we obtain}$$

$$\sum_{i=1}^{N}\sum_{j=1}^{q_i}\left(\log\Gamma(\alpha_{ij}) - \sum_{k=1}^{r_i}\log\Gamma(\alpha_{ijk})\right)$$
$$\geq \sum_{i=1}^{N}\alpha\log r_i - \frac{1}{2}\sum_{i=1}^{N} q_i(r_i-1)\log 2\pi$$
$$+ \frac{1}{2}\sum_{i=1}^{N}\sum_{j=1}^{q_i}\frac{(r_i-1)}{r_i}\sum_{k=1}^{r_i}\log\alpha_{ijk} + \mathcal{O}(1)$$
$$= \sum_{i=1}^{N}\alpha\log r_i + \frac{1}{2}\sum_{i=1}^{N} q_i(r_i-1)\log\frac{\alpha}{2\pi(r_i q_i)} + \mathcal{O}(1)$$

□

Theorem 3 shows also that when all the hyperparameters are greater than 1.0, the term $\frac{1}{2}\sum_{i=1}^{N} q_i(r_i-1)\log\frac{\alpha}{2\pi(r_i q_i)}$ dominates the prior term and increases with increasing $\alpha$. The increasing rate of the prior term decreases gradually as the number of parameters increases because of the term $\log\frac{\alpha}{2\pi(r_i q_i)}$. Consequently, the prior term produces complete graph structures when $\alpha$ becomes extremely large.

From Theorems 2 and 3, the prior term blocks the addition of arcs when $\alpha_{ijk} < 1.0$ for all the hyperparameters and helps to add arcs when $\alpha_{ijk} \geq 1.0$ for all hyperparameters. The role of the prior term changes rapidly from strongly blocking addition of arcs to strongly helping to add arcs dependent on $\alpha$. Therefore, the prior term is highly sensitive to $\alpha$ and might cause some odd phenomena of BDeu.

### 4.2 Analysis of likelihood term

This section presents analysis of the likelihood term in (6).

**Theorem 4.** *When $n + \alpha$ is sufficiently large, the likelihood term converges to*

$$\sum_{i=1}^{N}\sum_{j=1}^{q_i}\left(\sum_{k=1}^{r_i}\log\Gamma(\alpha_{ijk}+n_{ijk}) - \log\Gamma(\alpha_{ij}+n_{ij})\right)$$
$$= \log p(\widehat{\Theta} \mid \mathbf{X}, g, \alpha)$$
$$- \frac{1}{2}\sum_{i=1}^{N}\sum_{j=1}^{q_i}\left(\frac{(r_i-1)}{r_i}\sum_{k=1}^{r_i}\log(\frac{\alpha_{ijk}+n_{ijk}}{2\pi})\right) + \mathcal{O}(1).$$

*Proof.* When $n + \alpha$ is sufficiently large, using the Stirling series, we obtain

$$\sum_{i=1}^{N}\sum_{j=1}^{q_i}\left(\sum_{k=1}^{r_i}\log\Gamma(\alpha_{ijk}+n_{ijk}) - \log\Gamma(\alpha_{ij}+n_{ij})\right)$$
$$= \sum_{i=1}^{N}\sum_{j=1}^{q_i}\sum_{k=1}^{r_i}(\alpha_{ijk}+n_{ijk})\log\frac{(\alpha_{ijk}+n_{ijk})}{(\alpha_{ij}+n_{ij})}$$

$$+\frac{1}{2}\sum_{i=1}^{N}\sum_{j=1}^{q_i}\left((r_i-1)\log(2\pi)-\sum_{k=1}^{r_i}\log(\alpha_{ijk}+n_{ijk})\right.$$
$$\left.+\log(\alpha_{ij}+n_{ij})\right)+\mathcal{O}(max_i(\frac{r_iq_i}{n+\alpha}))$$

With Jensen's inequality,

$$\geq \sum_{i=1}^{N}\sum_{j=1}^{q_i}\sum_{k=1}^{r_i}(\alpha_{ijk}+n_{ijk})\log\frac{(\alpha_{ijk}+n_{ijk})}{(\alpha_{ij}+n_{ij})}$$
$$-\frac{1}{2}\sum_{i=1}^{N}\sum_{j=1}^{q_i}\left(\frac{(r_i-1)}{r_i}\sum_{k=1}^{r_i}\log(\frac{\alpha_{ijk}+n_{ijk}}{2\pi})\right)+\mathcal{O}(1).$$

□

Theorem 4 shows that the likelihood term can be decomposed into (1) the log posterior term $\sum_{i=1}^{N}\sum_{j=1}^{q_i}\sum_{k=1}^{r_i}(\alpha_{ijk}+n_{ijk})\log\frac{(\alpha_{ijk}+n_{ijk})}{(\alpha_{ij}+n_{ij})}$ and (2) the penalty term of complexity $\frac{1}{2}\sum_{i=1}^{N}\sum_{j=1}^{q_i}\left(\frac{(r_i-1)}{r_i}\sum_{k=1}^{r_i}\log(\frac{\alpha_{ijk}+n_{ijk}}{2\pi})\right)$.

The result shows that $\alpha_{ijk}$ in the likelihood term works as pseudo-data that augments data to support the hypothetical structure. Especially for BDeu, it works to avoid overfitting for parameter estimation regularization. In contrast, $\alpha$ in the prior term directly adds or deletes arcs according to the user's hypothetical structure $g^h$ for the learning structure regularization.

In addition, Steck and Jaakkola (2002) showed that asymptotically, as $\alpha$ approaches zero, the addition or deletion of an arc in a Bayesian network is infinitely favored or disfavored. The reason is explainable from the following theorem.

**Corollary 1.** *When $\alpha+n$ is sufficiently small, the likelihood term converges to*

$$\sum_{i=1}^{N}\sum_{j=1}^{q_i}\left(\sum_{k=1}^{r_i}\log\Gamma(\alpha_{ijk}+n_{ijk})-\log\Gamma(\alpha_{ij}+n_{ij})\right)$$
$$=-\sum_{i=1}^{N}q_i(r_i-1)\log\left(\frac{\alpha}{r_iq_i}\right)+\mathcal{O}(1).$$

*Proof.* When $\alpha+n<1.0$, then $n$ is expected to be zero. Therefore, when $n$ is zero and $\alpha$ approaches zero, we can prove the theorem in the same mode of Theorem 2. □

When using $\alpha\to 0$ and large data, BDeu yields an empty graph because the prior term dominates the BDeu, as shown in Theorems 2 and 4. However, when $\alpha\to 0$ and small data, if no zero sample cell exists in each $n_{ijk}$, then the prior term dominates the BDeu, as shown in Theorem 2. By increasing the number of arcs until some zero sample cells are generated, the change of the likelihood term becomes greater than that of the prior term. Consequently, BDeu adds extra arcs if zero sample cells are generated by the addition. Therefore, to solve this problem, determination of extremely small $\alpha$ for sparse data should be avoided.

### 4.3 Analysis of BDeu

Theorem 1(Ueno 2010) for BDeu holds only when $\alpha\geq r_iq_i$ and $n$ is large. Theorem 2, 3, and 4 relax the constraint $\alpha\geq r_iq_i$ to obtain the following more accurate result.

**Corollary 2.** *When $\alpha\geq r_iq_i, (i=1,\cdots,N)$ and $n$ is sufficiently large, log-BDeu converges to*

$$\log p(\boldsymbol{X}\mid g,\alpha)=\alpha\sum_{i=1}^{N}\log r_i+\log p(\widehat{\Theta}\mid \mathbf{X},g,\alpha) \quad (8)$$
$$-\frac{1}{2}\sum_{i=1}^{N}\sum_{j=1}^{q_i}\sum_{k=1}^{r_i}\frac{r_i-1}{r_i}\log\left(1+\frac{n_{ijk}}{\alpha_{ijk}}\right)+O(1),$$

*When $\alpha<r_iq_i, (i=1,\cdots,N)$ and $n$ is sufficiently large,*

$$\log p(\mathbf{X}\mid \mathbf{g},\alpha)=\log p(\widehat{\Theta}\mid \mathbf{X},g,\alpha) \quad (9)$$
$$-\frac{1}{2}\sum_{i=1}^{N}\sum_{j=1}^{q_i}\left(\frac{(r_i-1)}{r_i}\sum_{k=1}^{r_i}\log(\frac{n_{ijk}}{2\pi\alpha_{ijk}^2})\right)+\mathcal{O}(1),$$

*where $\alpha_{ijk}=\frac{\alpha}{r_iq_i}$.*

To be precise, the penalty term in (9) is $\log(\frac{\alpha_{ijk}+n_{ijk}}{2\pi\alpha_{ijk}^2})$ but $\alpha_{ijk}(<1.0)$ in the numerator is dropped because $n_{ijk}$ is sufficiently large. The result in (8) is the same as in Ueno (2010). However, the result in (9) is more important for BDeu than that in (8) because $\frac{\alpha}{r_iq_i}$ is usually less than 1.0 when the number of parameters becomes large. The main difference of the penalty terms in (8) and (9) is only the $\alpha_{ijk}$ in (8) and $\alpha_{ijk}^2$ in (9) because the constant term $2\pi$ can be dropped. Namely, term $\frac{1}{2}\sum_{i=1}^{N}\sum_{j=1}^{q_i}\left(\frac{(r_i-1)}{r_i}\sum_{k=1}^{r_i}\log(\frac{n_{ijk}}{2\pi\alpha_{ijk}^2})\right)$ in (9) also reflects the difference between the learned structure from data and the hypothetical structure $g^h$ from the user's knowledge. From (8), when $\alpha_{ijk}\geq 1$, the penalty term with large $\alpha_{ijk}$ decreases to add the arcs which are included in the user's hypothetical structures. From (9), when $\alpha_{ijk}<1$, the penalty term with small $\alpha_{ijk}$ increases to delete the arcs which are not included in the user's hypothetical structures.

For BDeu, $\alpha_{ijk}^2=(\frac{\alpha}{r_iq_i})^2, (j=1,\cdots,q_i,k=1,\cdots,r_i)$ in (9) also takes the same value for all the parameters given the number of parameters. Therefore, the penalty term of BDeu with small $\alpha$ also reflects the difference between the empirical distri-bution $n_{ijk}$ and the uniform distribution of $\alpha_{ijk}$. Moreover, the penalty magnitude of complexity in (9) rapidly becomes larger as the number of parameters increases compared with that in (8) because of the term $\alpha_{ijk}^2$. Consequently, BDeu has the complexity penalty term that increases as the empirical distribution becomes skewed (non-uniform).

Regarding sensitivity problems, they are explainable as follows. The optimal $\alpha$ should become large to increase the magnitude of the penalty term when the empirical distribution is uniform because the user's hypothetical structure (the uniform distribution of $\alpha_{ijk}$) is true. In contrast, the optimal $\alpha$ should become small when the empirical distribution is skewed because the user's hypothetical structure is not true. Consequently, the uniform distribution assumption in BDeu suffers from the sensitivity of BDeu to $\alpha$. Steck (2008) reported that the optimal $\alpha$ of BDeu becomes small when the conditional distributions of the variables are very skewed. Our analysis agrees with this result.

Next, we will consider the meaning of the prior in BDeu. From Theorems 2, 3 and 4, only the prior term reflects $\alpha_{ijk}$ in the penalty terms in (8) and (9). Namely, the prior term represents the user's hypothetical structure $g^h$ in the penalty term which reflects the difference between the empirical distribution and the prior distribution. Accordingly, the prior term is extremely important when we use the user's approximate prior knowledge for BDe. However, when we use BDeu with no prior knowledge, the prior term might suffer some unstable or undesirable results because of the sensitivity. The reason is that the prior term does not represent ignorance of prior knowledge but a user's prior belief for the uniformity of conditional distribution. Therefore, when the empirical distribution is skewed, BDeu can be expected to reduce the degree of the prior belief by reducing the $\alpha$ value because the hypothetical structure is not true. In contrast, when the empirical distribution is uniform, BDeu can be expected to increase the degree of the prior belief by increasing the $\alpha$ value because the hypothetical structure is true. Consequently, BDeu does not assume a non-informative prior but rather a user's prior belief for the uniformity of conditional distribution. Therefore, empirical approaches (Silander, Kontkanen, and Myllymaki, 2007, Steck 2008) from actual data are useful to adjust the degree of prior belief. However, these approaches leave the following problem. Actually, the degrees of uniformity for the variables differ greatly although the only one optimum $\alpha$ of BDeu for a network is determined from the uniformities of all variables. Therefore, these methods make it difficult to find an approximate $\alpha$ for learning a network that has combined skewed and uniform conditional distributions. In addition, Bayesian networks usually have combined skewed and uniform distributions.

## 5 Robust learning for prior belief

The previous section identified the factors that might cause the sensitivity of BDeu. However, it is impossible to eliminate the penalty term for the skewness (non-uniformity) of the conditional distribution directly from BDeu because it can not be decomposed from log-BDeu. Therefore, we eliminate the sensitive factors from the approximation in Corollary 2. The sensitive factors in the approximation are the penalty term for the skewness of the conditional distribution in (8) and (9).

First, we replace the prior term of BDeu by a constant term because it is highly sensitive to $\alpha$. This procedure is equivalent to ignore the prior term. Then we consider only the likelihood term. The likelihood term still has the difference penalty for the skewness of conditional distribution because $n_{ijk}$ remains in the penalty term. For that reason, we replace the likelihood term by the lower bound. The following theorem to obtain the lower bound is provable from Theorem 4.

**Corollary 3.** *When $\alpha + n$ is sufficiently large, the lower bound of the likelihood term can be approximated as*

$$\sum_{i=1}^{N}\sum_{j=1}^{q_i}\left(\sum_{k=1}^{r_i}\log\Gamma(\alpha_{ijk}+n_{ijk})-\log\Gamma(\alpha_{ij}+n_{ij})\right)$$
$$\geq \log p(\widehat{\Theta}\mid \mathbf{X},g,\alpha)-\frac{1}{2}k(g)\log(\alpha+n), \qquad (10)$$

*where $k(g)$ is the number of parameters and $\alpha_{ijk}=\dfrac{\alpha}{r_iq_i}$.*

*Proof.* From Theorem 4,
$$\sum_{i=1}^{N}\sum_{j=1}^{q_i}\left(\sum_{k=1}^{r_i}\log\Gamma(\alpha_{ijk}+n_{ijk})-\log\Gamma(\alpha_{ij}+n_{ij})\right)$$
$$\geq \log p(\widehat{\Theta}\mid \mathbf{X},g,\alpha)$$
$$-\frac{1}{2}\sum_{i=1}^{N}\sum_{j=1}^{q_i}\left(\frac{(r_i-1)}{r_i}\sum_{k=1}^{r_i}\log(\frac{\alpha_{ijk}+n_{ijk}}{2\pi})\right)$$
because $\alpha\geq\alpha_{ijk}, n\geq n_{ijk}$ for $\forall i,\forall j,\forall k$,
$$\geq \log p(\widehat{\Theta}\mid \mathbf{X},g,\alpha)-\frac{1}{2}k(g)\log\frac{(\alpha+n)}{2\pi}.$$

We obtain (10) because $2\pi$ is dropped when $\alpha+n$ is large. □

This paper presents a proposal of a learning score for Bayesian networks by maximizing the right side of (10). We designate the score in (10) "non-informative prior Bayesian information criterion (NIP-BIC)".

NIP-BIC is similar to the Bayesian information criterion (BIC;Schwarz,1978) in (11) because NIP-BIC converges asymptotically to BIC for sufficiently large data.

$$BIC = \sum_{i=1}^{N} \sum_{j=1}^{q_i} \sum_{k=1}^{r_i} n_{ijk} \log \frac{n_{ijk}}{n_{ij}} - \frac{1}{2} k(g) \log n. \quad (11)$$

Suzuki (1993) and Bouckaert (1994) proposed a learning Bayesian network using BIC based on the minimum description length (MDL) principle. The most important difference between the BIC and NIP-BIC is that the hyper-parameters remain in the term $\sum_{k=0}^{r_i-1}(\alpha_{ijk}+n_{ijk})\log\frac{(\alpha_{ijk}+n_{ijk})}{(\alpha_{ij}+n_{ij})}$ in (10), but do not remain in BIC. When the data are few, learning Bayesian networks with the BIC is well known to tend to overfit the data. In fact, some studies (e.g. Yang 2002) have demonstrated that the performance of the BIC for Bayesian networks is somewhat unstable when insufficient data are given. Because the likelihood $\sum_{i=1}^{N}\sum_{j=1}^{q_i}\sum_{k=1}^{r_i} n_{ijk}\log\frac{n_{ijk}}{n_{ij}}$ is very sensitive to variation in $n_{ijk}$, and because $n_{ijk}$ might vary with the database size, increasing the amount of data might not reduce the chance of making errors during structure induction. Therefore, the role of $\alpha$ in the log-posterior term in (10) is important.

Additionally, BDeu when $\alpha = 1.0$ is mostly approximated by NIP-BIC because $r_i q_i n_{ijk}$ in the penalty term of (8) and (9) can be approximated by $n$ since $E_{jk}[r_i q_i n_{ijk}] = n$. Ueno (2010) recommended the BDeu with $\alpha = 1.0$ corresponding to the smallest positive assignment of the hyperparameters, which allows the data to reflect the estimated parameters to the greatest degree possible. Additionally, the variances of the Dirichlet distribution are known to decrease with the sum of the hyperparameters (Castillo, Hadi, and Solares, 1997). This decrease of variance also suggests that BDeu with $\alpha = 1.0$ is the best method to mitigate the influence of ESS for parameter estimation. This paper therefore also recommends $\alpha = 1.0$ as the ESS of NIP-BIC. Results show that BDeu with $\alpha = 1.0$ assumes the uniformity of conditional distribution and remains sensitive to $\alpha$ because it remains $n_{ijk}$ in the penalty term.

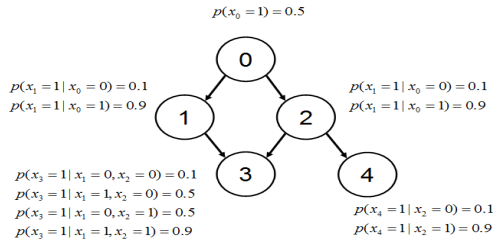

Figure 1: g1: Strongly skewed distribution

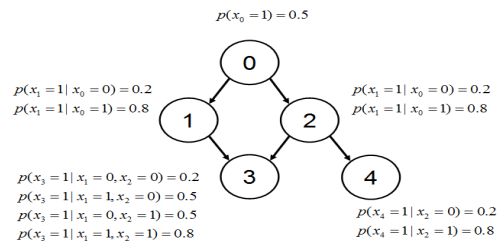

Figure 2: g2: Skewed distribution

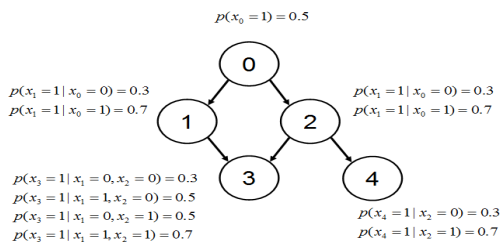

Figure 3: g3: Uniform distribution

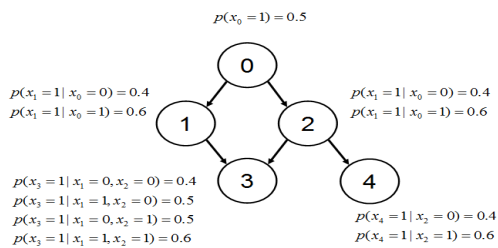

Figure 4: g4: Strongly uniform distribution

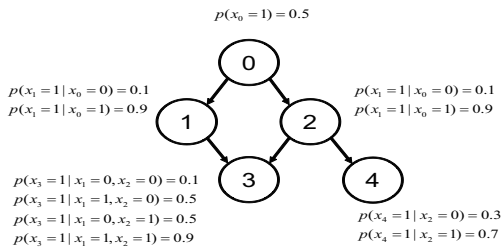

Figure 5: g5: Combined skewed and uniform distributions

## 6 Numerical examples

We conducted simulation experiments to compare NIP-BIC with BDeu. We used small network struc-

Table 1: Learning with ML and BDeu by changing $\alpha$

| g1 | ML($\alpha_{ijk}=1.0$) | | | BDeu($\alpha=0.01$) | | | BDeu($\alpha=0.1$) | | | BDeu($\alpha=1.0$) | | | BDeu($\alpha=10$) | | | BDeu($\alpha=100$) | | |
|---|---|---|---|---|---|---|---|---|---|---|---|---|---|---|---|---|---|---|
| n | | + | - | | + | - | | + | - | | + | - | | + | - | | + | - |
| 50 | 19 | 67 | 143 | 3 | 52 | 131 | 5 | 51 | 125 | 20 | 58 | 96 | 8 | 204 | 33 | 0 | 491 | 0 |
| 100 | 69 | 31 | 19 | 4 | 25 | 114 | 8 | 26 | 110 | 32 | 31 | 69 | 21 | 139 | 14 | 0 | 486 | 0 |
| 200 | 77 | 23 | 2 | 3 | 18 | 109 | 41 | 14 | 67 | 84 | 9 | 11 | 50 | 58 | 2 | 0 | 467 | 0 |
| 500 | 54 | 50 | 0 | 94 | 3 | 3 | 97 | 3 | 0 | 92 | 8 | 0 | 74 | 30 | 0 | 0 | 433 | 0 |
| 1000 | 39 | 81 | 0 | 100 | 0 | 0 | 100 | 0 | 0 | 99 | 1 | 0 | 87 | 14 | 0 | 0 | 409 | 0 |

| g2 | ML($\alpha_{ijk}=1.0$) | | | BDeu($\alpha=0.01$) | | | BDeu($\alpha=0.1$) | | | BDeu($\alpha=1.0$) | | | BDeu($\alpha=10$) | | | BDeu($\alpha=100$) | | |
|---|---|---|---|---|---|---|---|---|---|---|---|---|---|---|---|---|---|---|
| n | | + | - | | + | - | | + | - | | + | - | | + | - | | + | - |
| 50 | 31 | 44 | 75 | 1 | 16 | 228 | 2 | 25 | 163 | 4 | 37 | 125 | 19 | 95 | 63 | 0 | 354 | 9 |
| 100 | 39 | 73 | 30 | 0 | 18 | 129 | 0 | 21 | 121 | 12 | 27 | 103 | 39 | 68 | 44 | 0 | 368 | 9 |
| 200 | 31 | 90 | 4 | 1 | 14 | 113 | 12 | 13 | 100 | 50 | 16 | 52 | 65 | 38 | 8 | 0 | 340 | 1 |
| 500 | 34 | 107 | 0 | 66 | 1 | 35 | 95 | 0 | 5 | 99 | 1 | 0 | 89 | 11 | 0 | 1 | 262 | 0 |
| 1000 | 18 | 143 | 0 | 100 | 0 | 0 | 100 | 0 | 0 | 100 | 0 | 0 | 92 | 8 | 0 | 16 | 157 | 0 |

| g3 | ML($\alpha_{ijk}=1.0$) | | | BDeu($\alpha=0.01$) | | | BDeu($\alpha=0.1$) | | | BDeu($\alpha=1.0$) | | | BDeu($\alpha=10$) | | | BDeu($\alpha=100$) | | |
|---|---|---|---|---|---|---|---|---|---|---|---|---|---|---|---|---|---|---|
| n | | + | - | | + | - | | + | - | | + | - | | + | - | | + | - |
| 50 | 9 | 66 | 115 | 0 | 2 | 421 | 0 | 9 | 349 | 0 | 31 | 218 | 5 | 78 | 139 | 1 | 232 | 63 |
| 100 | 14 | 92 | 64 | 0 | 1 | 309 | 0 | 3 | 215 | 0 | 14 | 149 | 9 | 57 | 104 | 7 | 186 | 44 |
| 200 | 19 | 123 | 22 | 0 | 1 | 168 | 0 | 5 | 131 | 4 | 6 | 105 | 31 | 30 | 58 | 13 | 156 | 20 |
| 500 | 17 | 146 | 3 | 2 | 1 | 101 | 11 | 2 | 90 | 52 | 3 | 48 | 77 | 17 | 11 | 39 | 92 | 4 |
| 1000 | 18 | 152 | 0 | 47 | 0 | 53 | 83 | 0 | 17 | 97 | 0 | 3 | 99 | 1 | 0 | 69 | 36 | 0 |

| g4 | ML($\alpha_{ijk}=1.0$) | | | BDeu($\alpha=0.01$) | | | BDeu($\alpha=0.1$) | | | BDeu($\alpha=1.0$) | | | BDeu($\alpha=10$) | | | BDeu($\alpha=100$) | | |
|---|---|---|---|---|---|---|---|---|---|---|---|---|---|---|---|---|---|---|
| n | | + | - | | + | - | | + | - | | + | - | | + | - | | + | - |
| 50 | 2 | 130 | 243 | 0 | 1 | 497 | 0 | 5 | 477 | 0 | 17 | 421 | 1 | 85 | 302 | 2 | 209 | 192 |
| 100 | 6 | 143 | 154 | 0 | 1 | 490 | 0 | 4 | 457 | 0 | 16 | 364 | 0 | 53 | 248 | 4 | 171 | 138 |
| 200 | 5 | 146 | 87 | 0 | 0 | 458 | 0 | 1 | 399 | 0 | 4 | 281 | 4 | 25 | 185 | 5 | 119 | 104 |
| 500 | 17 | 135 | 37 | 0 | 0 | 274 | 0 | 0 | 202 | 2 | 1 | 144 | 13 | 9 | 97 | 25 | 65 | 57 |
| 1000 | 19 | 148 | 12 | 0 | 0 | 168 | 1 | 0 | 139 | 8 | 0 | 107 | 29 | 5 | 75 | 51 | 36 | 30 |

| g5 | ML($\alpha_{ijk}=1.0$) | | | BDeu($\alpha=0.01$) | | | BDeu($\alpha=0.1$) | | | BDeu($\alpha=1.0$) | | | BDeu($\alpha=10$) | | | BDeu($\alpha=100$) | | |
|---|---|---|---|---|---|---|---|---|---|---|---|---|---|---|---|---|---|---|
| n | | + | - | | + | - | | + | - | | + | - | | + | - | | + | - |
| 50 | 11 | 68 | 158 | 0 | 41 | 211 | 3 | 51 | 184 | 7 | 69 | 149 | 4 | 162 | 72 | 0 | 363 | 15 |
| 100 | 45 | 63 | 53 | 1 | 29 | 156 | 2 | 34 | 138 | 27 | 46 | 97 | 23 | 114 | 37 | 0 | 368 | 0 |
| 200 | 48 | 59 | 6 | 10 | 26 | 116 | 37 | 23 | 78 | 80 | 13 | 21 | 47 | 67 | 7 | 0 | 358 | 0 |
| 500 | 42 | 77 | 0 | 96 | 0 | 4 | 99 | 0 | 1 | 96 | 4 | 0 | 80 | 21 | 0 | 0 | 324 | 0 |
| 1000 | 25 | 102 | 0 | 100 | 0 | 0 | 99 | 1 | 0 | 99 | 1 | 0 | 89 | 13 | 0 | 0 | 281 | 0 |

Table 2: Learning with BIC and NIP-BIC by changing $\alpha$

| g1 | BIC | | | NIP($\alpha=0.01$) | | | NIP($\alpha=0.1$) | | | NIP($\alpha=1.0$) | | | NIP($\alpha=10$) | | | NIP($\alpha=100$) | | |
|---|---|---|---|---|---|---|---|---|---|---|---|---|---|---|---|---|---|---|
| n | | + | - | | + | - | | + | - | | + | - | | + | - | | + | - |
| 50 | 0 | 476 | 7 | 18 | 46 | 117 | 18 | 45 | 118 | 15 | 41 | 121 | 7 | 41 | 132 | 0 | 40 | 147 |
| 100 | 0 | 452 | 7 | 48 | 14 | 61 | 48 | 14 | 61 | 49 | 14 | 60 | 49 | 16 | 58 | 21 | 43 | 78 |
| 200 | 0 | 398 | 3 | 91 | 5 | 8 | 91 | 5 | 8 | 92 | 4 | 7 | 95 | 3 | 5 | 20 | 117 | 8 |
| 500 | 1 | 302 | 0 | 99 | 1 | 0 | 99 | 1 | 0 | 100 | 0 | 0 | 99 | 1 | 0 | 13 | 127 | 0 |
| 1000 | 27 | 145 | 0 | 100 | 0 | 0 | 100 | 0 | 0 | 100 | 0 | 0 | 100 | 0 | 0 | 39 | 79 | 0 |

| g2 | BIC | | | NIP($\alpha=0.01$) | | | NIP($\alpha=0.1$) | | | NIP($\alpha=1.0$) | | | NIP($\alpha=10$) | | | NIP($\alpha=100$) | | |
|---|---|---|---|---|---|---|---|---|---|---|---|---|---|---|---|---|---|---|
| n | | + | - | | + | - | | + | - | | + | - | | + | - | | + | - |
| 50 | 0 | 473 | 9 | 8 | 29 | 115 | 8 | 29 | 116 | 6 | 26 | 118 | 2 | 22 | 128 | 0 | 7 | 278 |
| 100 | 0 | 422 | 11 | 24 | 26 | 89 | 24 | 26 | 89 | 22 | 26 | 91 | 21 | 21 | 96 | 0 | 17 | 121 |
| 200 | 1 | 323 | 2 | 67 | 12 | 35 | 67 | 12 | 35 | 67 | 12 | 35 | 68 | 9 | 36 | 53 | 9 | 52 |
| 500 | 55 | 75 | 0 | 98 | 2 | 0 | 98 | 2 | 0 | 98 | 2 | 0 | 99 | 1 | 0 | 96 | 4 | 0 |
| 1000 | 81 | 22 | 0 | 100 | 0 | 0 | 100 | 0 | 0 | 100 | 0 | 0 | 100 | 0 | 0 | 100 | 0 | 0 |

| g3 | BIC | | | NIP($\alpha=0.01$) | | | NIP($\alpha=0.1$) | | | NIP($\alpha=1.0$) | | | NIP($\alpha=10$) | | | NIP($\alpha=100$) | | |
|---|---|---|---|---|---|---|---|---|---|---|---|---|---|---|---|---|---|---|
| n | | + | - | | + | - | | + | - | | + | - | | + | - | | + | - |
| 50 | 0 | 446 | 24 | 0 | 33 | 191 | 0 | 33 | 191 | 0 | 26 | 202 | 0 | 18 | 225 | 0 | 0 | 448 |
| 100 | 0 | 346 | 47 | 2 | 19 | 135 | 2 | 19 | 135 | 1 | 18 | 138 | 0 | 10 | 145 | 0 | 3 | 248 |
| 200 | 21 | 138 | 40 | 11 | 8 | 97 | 11 | 8 | 97 | 11 | 8 | 97 | 9 | 6 | 99 | 3 | 6 | 114 |
| 500 | 67 | 34 | 6 | 70 | 3 | 29 | 70 | 3 | 29 | 70 | 3 | 29 | 69 | 3 | 30 | 60 | 2 | 40 |
| 1000 | 90 | 11 | 0 | 99 | 0 | 1 | 99 | 0 | 1 | 99 | 0 | 1 | 99 | 0 | 1 | 99 | 0 | 1 |

| g4 | BIC | | | NIP($\alpha=0.01$) | | | NIP($\alpha=0.1$) | | | NIP($\alpha=1.0$) | | | NIP($\alpha=10$) | | | NIP($\alpha=100$) | | |
|---|---|---|---|---|---|---|---|---|---|---|---|---|---|---|---|---|---|---|
| n | | + | - | | + | - | | + | - | | + | - | | + | - | | + | - |
| 50 | 0 | 427 | 99 | 0 | 26 | 400 | 0 | 26 | 400 | 0 | 25 | 407 | 0 | 14 | 431 | 0 | 1 | 498 |
| 100 | 3 | 282 | 143 | 0 | 21 | 332 | 0 | 21 | 332 | 0 | 21 | 333 | 0 | 18 | 359 | 0 | 2 | 473 |
| 200 | 6 | 75 | 145 | 0 | 7 | 250 | 0 | 7 | 250 | 0 | 7 | 252 | 0 | 7 | 257 | 0 | 3 | 351 |
| 500 | 23 | 17 | 83 | 6 | 3 | 122 | 6 | 3 | 122 | 6 | 3 | 122 | 5 | 3 | 125 | 2 | 1 | 149 |
| 1000 | 36 | 7 | 64 | 13 | 1 | 100 | 13 | 1 | 100 | 13 | 1 | 100 | 13 | 1 | 100 | 10 | 0 | 105 |

| g5 | BIC | | | NIP($\alpha=0.01$) | | | NIP($\alpha=0.1$) | | | NIP($\alpha=1.0$) | | | NIP($\alpha=10$) | | | NIP($\alpha=100$) | | |
|---|---|---|---|---|---|---|---|---|---|---|---|---|---|---|---|---|---|---|
| n | | + | - | | + | - | | + | - | | + | - | | + | - | | + | - |
| 50 | 0 | 459 | 21 | 8 | 52 | 149 | 8 | 52 | 150 | 6 | 50 | 153 | 5 | 50 | 162 | 0 | 26 | 227 |
| 100 | 0 | 429 | 10 | 34 | 32 | 88 | 34 | 32 | 87 | 34 | 32 | 87 | 37 | 34 | 82 | 18 | 45 | 111 |
| 200 | 3 | 276 | 5 | 82 | 13 | 17 | 82 | 13 | 17 | 82 | 13 | 17 | 86 | 10 | 14 | 40 | 64 | 18 |
| 500 | 32 | 94 | 0 | 100 | 0 | 0 | 100 | 0 | 0 | 100 | 0 | 0 | 100 | 0 | 0 | 38 | 63 | 0 |
| 1000 | 66 | 43 | 0 | 100 | 0 | 0 | 100 | 0 | 0 | 100 | 0 | 0 | 100 | 0 | 0 | 61 | 39 | 0 |

tures with binary variables in Figs. 1, 2, 3, and 4, in which the distributions are changed from skewed to uniform. Figure 1 presents a structure in which the conditional probabilities differ greatly because of the parent variable states (g1: Strongly skewed distribution). By gradually reducing the difference of the conditional probabilities from Fig. 1, we generated in Fig. 2(g2: Skewed distribution), Fig. 3(g3:

Uniform distribution), and Fig. 4(g4: Strongly uniform distribution). Additionally, we generated in Fig. 5(g5:Combined skewed and uniform distributions) which has combined skewed and uniform conditional distributions.

Procedures used for the simulation experiments are described below.

1. We generated 50, 100, 200, 500, and 1,000 samples from the five figures.

2. Using the marginal likelihood(ML)($\alpha_{ijk} = 1.0$)(Cooper and Herskovits, 1992), BIC, BDeu and NIP-BIC by changing $\alpha$ (0.01, 0.1, 1.0, 10, 100), Bayesian network structures were estimated, respectively, based on 50, 100, 200, 500, and 1,000 samples. We searched for the true structure among all possible structures.

3. The times the estimated structure was the true structure were counted by repeating procedure 2 for 100 iterations.

Table 1 presents the results for ML and BDeu. Table 2 presents the results for BIC and NIP-BIC. Column "+"has the total number of extra arcs for the estimated structures, and column "-" has the total number of missing arcs for the estimated structures. The maximum quantities of "+" and "-" are both 500. Column "O" has the number of correct-structure estimates in 100 trials.

The results presented in Table 1 reveal that BDeu is highly sensitive to $\alpha$. That is, the optimum value of $\alpha$ becomes small as the conditional distribution becomes skewed. In contrast, the optimum value of $\alpha$ becomes large as the conditional distribution becomes uniform. As described in section 4, the uniformity of conditional distribution as the user's prior belief causes this sensitivity. The ML with $\alpha_{ijk} = 1.0$, which also assumes the prior uniformity with no constraints of the number of parameters, shows much less accurate performance because it does not satisfy the likelihood equivalence.

As shown in Table 2, the performance of NIP-BIC is better than that of BDeu in almost all cases except g4 ($\alpha = 10$ and 100). Actually, BDeu provides better performance when the conditional distribution becomes uniform and when $\alpha$ becomes large because the uniformity of conditional distribution is assumed as discussed in section 4. Especially, for g5, NIP-BIC strongly improves performance of BDeu for all $\alpha$ settings. The reason is that learning a network combining skewed and uniform distributions is very difficult to set an approximate $\alpha$ for BDeu. This advantage is the most important for NIP-BIC because Bayesian networks usually have combined skewed and uniform distributions. The results also show that the performance of BDeu is highly sensitive to $\alpha$ but the performance of NIP-BIC is robust for $\alpha$.

Additionally, it is noteworthy that the NIP-BIC performs similarly to BDeu with $\alpha = 1.0$ because BDeu with $\alpha = 1.0$ is mostly approximated by NIP-BIC, as explained in Section 5. The results also show that NIP-BIC provides better performance than BDeu with $\alpha = 1.0$ because BDeu assumes the uniformity of conditional distribution and because it remains sensitive to $\alpha$.

The results for BIC show a lower level of performance than the others because BIC tends to overfit the data. The results also indicate that NIP-BIC tends to suffer more missing arcs than BIC. The reason is that the hyper-parameters remain in the term of NIP-BIC $\sum_{k=0}^{r_i-1}(\alpha_{ijk} + n_{ijk})\log \frac{(\alpha_{ijk}+n_{ijk})}{(\alpha_{ij}+n_{ij})}$ in (10), but they do not remain in BIC. The hyper-parameters work to block overfitting for large data but tend to suffer missing arcs for small data.

Consequently, NIP-BIC is a convenient learning score when given no prior knowledge. As a result, we can confirm that NIP-BIC relaxes the sensitivity of BDeu to $\alpha$ and that it yields more robust estimators.

# 7 Conclusions

This paper presented asymptotic analysis of log-BDeu by improving the result reported by Ueno (2010). Although the final asymptotic result was almost identical to that presented earlier by Ueno (2010), we also newly discovered that the penalty term rapidly becomes larger, especially when the hyperparameters are less than 1.0. Furthermore, we identified the reasons of the sensitivity of BDeu using some asymptotic analyses by decomposing it into the following two parts: (1) a prior term that is independent of data, and (2) a likelihood term that reflects data. Moreover, results show that the prior term changes rapidly the role from strongly blocking extra arcs to strongly helping additional arcs. The results show that the prior term acts as to be highly sensitive to ESS and that it causes some odd phenomena of BDeu. The results also showed that the prior of BDeu does not represent ignorance of prior knowledge but rather a user's prior belief in the uniformity of conditional distribution. The results further imply that the optimal ESS becomes large/small when the empirical conditional distribution becomes uniform/skewed. This main factor underpins the sensitivity of BDeu to ESS. Moreover, to solve the sensitivity problem, we proposed a robust learning score (called "NIP-BIC") for ESS by eliminating the sensitivity factors from the approximation of log-BDeu because it is impossible to eliminate them directly from the log-BDeu function. Some numerical experiments have shown that NIP-BIC is effective, especially when we have no prior knowledge.

This study elucidated the learning performances of NIP-BIC using small network structures. Finding the MAP estimate of the structure is an NP-complete problem (Chickering, 1996). Recently however, the exact solution methods can produce results in reasonable computation time if the variables are not prohibitively numerous (ex. Silander and Myllymaki, 2006, Perrier *et al.*, 2008). An important future task is evaluation of the performances of NIP-BIC for large network structures from various perspectives.